\documentclass[11pt,a4paper,hyphens]{article}
\usepackage[hyperref]{acl2019}
\usepackage{times}
\usepackage{latexsym,graphicx,amssymb,multirow,amsmath,booktabs}
\usepackage{balance}
\usepackage{flushend}
\usepackage{soul}
\usepackage{subfigure}
\usepackage[font={footnotesize}]{caption}
\usepackage{url}
\usepackage{algorithm}
\usepackage{algpseudocode}
\newcommand{\ModelName}{{\textsc{Capsule-NLU}}}

\newcommand{\FirstCapsule}{WordCaps}
\newcommand{\SecondCapsule}{SlotCaps}
\newcommand{\ThirdCapsule}{IntentCaps}

\newcommand\blfootnote[1]{%
  \begingroup
  \renewcommand\thefootnote{}\footnote{#1}%
  \addtocounter{footnote}{-1}%
  \endgroup
}

\aclfinalcopy
\title{Joint Slot Filling and Intent Detection via Capsule Neural Networks} 

\author{Chenwei Zhang\footnotemark[2]~,~Yaliang Li\footnotemark[4]~,~Nan Du\footnotemark[3]~,~Wei Fan\footnotemark[3]~,~Philip S. Yu\footnotemark[2]~\footnotemark[5]\\
  \footnotemark[2]~University of Illinois at Chicago, Chicago, IL 60607 USA\\
  \footnotemark[4]~Alibaba Group, Bellevue, WA 98004 USA\\
  \footnotemark[3]~Tencent Medical AI Lab, Palo Alto, CA 94301 USA\\
  \footnotemark[5]~~Institute for Data Science, Tsinghua University, Beijing, China\\
  {\tt \{czhang99,psyu\}@uic.edu}, {\tt yaliang.li@alibaba-inc.com},\\
  {\tt nandu2048@gmail.com}, {\tt davidwfan@tencent.com}\\
  }
\date{}

\begin{document}
\maketitle
\begin{abstract}
Being able to recognize words as slots and detect the intent of an utterance has been a keen issue in natural language understanding. The existing works either treat slot filling and intent detection separately in a pipeline manner, or adopt joint models which sequentially label slots while summarizing the utterance-level intent without explicitly preserving the hierarchical relationship among words, slots, and intents. To exploit the semantic hierarchy for effective modeling, we propose a capsule-based neural network model which accomplishes slot filling and intent detection via a dynamic routing-by-agreement schema. A re-routing schema is proposed to further synergize the slot filling performance using the inferred intent representation. Experiments on two real-world datasets show the effectiveness of our model when compared with other alternative model architectures, as well as existing natural language understanding services\blfootnote{Code and data available at \url{https://github.com/czhang99/Capsule-NLU}}.
\end{abstract}

\section{Introduction}
\begin{figure}[htbp]
    \centering
    \includegraphics[width=\linewidth]{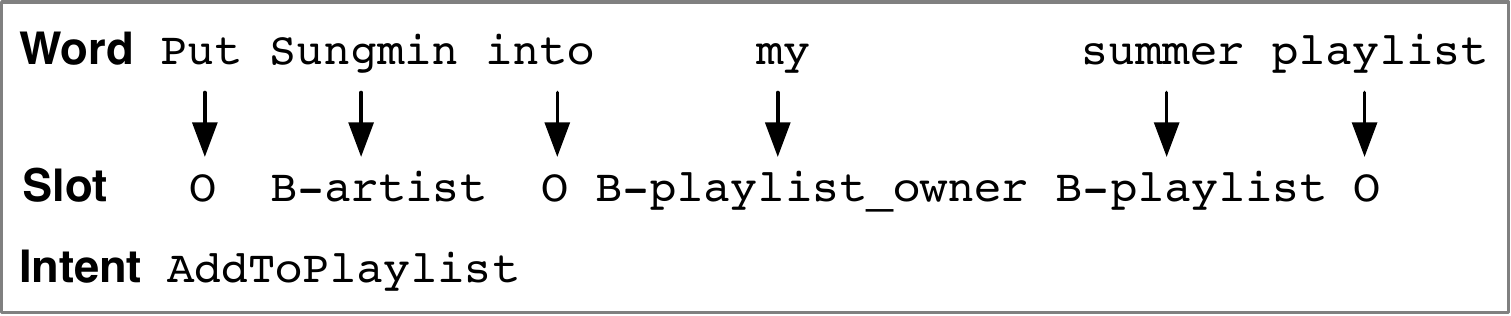}
    \vspace{0.01in}
    \caption{An example of an utterance with BOI format annotation for slot filling, which indicates the slot of artist, play list owner, and play list name from an utterance with an intent AddToPlaylist.}\label{fig::illustration}
\end{figure}

With the ever-increasing accuracy in speech recognition and complexity in user-generated utterances, it becomes a critical issue for mobile phones or smart speaker devices to understand the natural language in order to give informative responses. Slot filling and intent detection play important roles in Natural Language Understanding (NLU) systems.
For example, given an utterance from the user, the slot filling annotates the utterance on a word-level, indicating the slot type mentioned by a certain word such as the slot \texttt{artist} mentioned by the word \texttt{Sungmin}, while the intent detection works on the utterance-level to give categorical intent label(s) to the whole utterance. Figure \ref{fig::illustration} illustrates this idea.

\begin{figure*}[htbp]
    \centering
    \includegraphics[width=\linewidth]{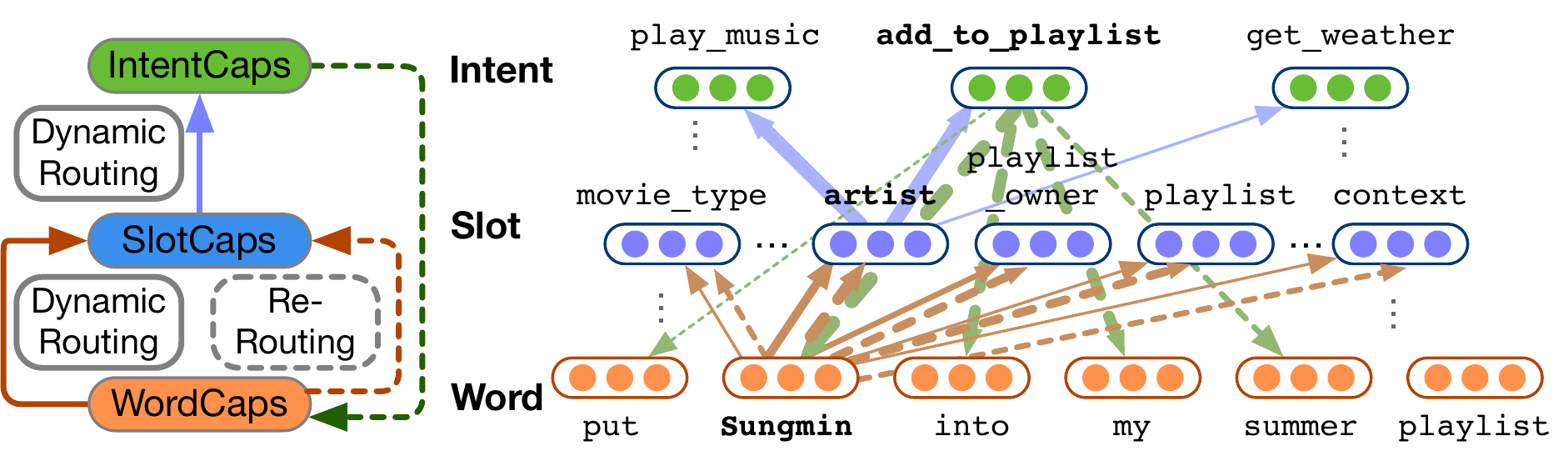}
    \vspace{0.05in}
    \caption{Illustration of the proposed {\ModelName} model for joint slot filling and intent detection. The model does slot filling by learning to assign each word in the {\FirstCapsule} to the most appropriate slot in {\SecondCapsule} via dynamic routing. The weights learned via dynamic routing indicate how strong each word in {\FirstCapsule} belongs to a certain slot type in {\SecondCapsule}. The dynamic routing also learns slot representations using {\FirstCapsule} and the learned weight. The learned slot representations in {\SecondCapsule} are further aggregated to predict the utterance-level intent of the utterance. Once the intent label of the utterance is determined, a novel re-routing process is proposed to help improve word-level slot filling by the inferred utterance-level intent label. The solid lines indicate the dynamic-routing process and dash lines indicate the re-routing process. 
    }
    \label{fig::overall}
\end{figure*}

To deal with diversely expressed utterances without additional feature engineering, deep neural network based user intent detection models \citep{hu2009understanding,xu2013convolutional,zhang2016mining,liu2016attention,zhang2017bringing,chen2016end,xia2018zero} are proposed to classify user intents given their utterances in the natural language. 

Currently, the slot filling is usually treated as a sequential labeling task. A neural network such as a recurrent neural network (RNN) or a convolution neural network (CNN) is used to learn context-aware word representations, along with sequence tagging methods such as conditional random field (CRF) \citep{lafferty2001conditional} that infer the slot type for each word in the utterance.

Word-level slot filling and utterance-level intent detection can be conducted simultaneously to achieve a synergistic effect.
The recognized slots, which possess word-level signals, may give clues to the utterance-level intent of an utterance. For example, with a word \texttt{Sungmin} being recognized as a slot \texttt{artist}, the utterance is more likely to have an intent of \texttt{AddToPlayList} than other intents such as \texttt{GetWeather} or \texttt{BookRestaurant}. 

Some existing works learn to fill slots while detecting the intent of the utterance \cite{xu2013convolutional,hakkani2016multi,liu2016attention,goo2018slot}: a convolution layer or a recurrent layer is adopted to sequentially label word with their slot types: the last hidden state of the recurrent neural network, or an attention-weighted sum of all convolution outputs are used to train an utterance-level classification module for intent detection. Such approaches achieve decent performances but do not explicitly consider the hierarchical relationship between words, slots, and intents: intents are sequentially summarized from the word sequence.
As the sequence becomes longer, it is risky to simply rely on the gate function of RNN to compress all context information in a single vector \cite{cheng2016long}.

In this work, we make the very first attempt to bridge the gap between word-level slot modeling and the utterance-level intent modeling via a hierarchical capsule neural network structure \citep{hinton2011transforming,sabour2017dynamic}.
A capsule houses a vector representation of a group of neurons.
The capsule model learns a hierarchy of feature detectors via a routing-by-agreement mechanism: capsules for detecting low-level features send their outputs to high-level capsules only when there is a strong agreement of their predictions to high-level capsules.

The aforementioned properties of capsule models are appealing for natural language understanding from a hierarchical perspective: words such as \texttt{Sungmin} are routed to concept-level slots such as \texttt{artist}, by learning how each word matches the slot representation. Concept-level slot features such as \texttt{artist}, \texttt{playlist owner}, and \texttt{playlist} collectively contribute to an utterance-level intent \texttt{AddToPlaylist}. The dynamic routing-by-agreement assigns a larger weight from a lower-level capsule to a higher-level when the low-level feature is more predictive to one high-level feature, than other high-level features. Figure \ref{fig::overall} illustrates this idea.

The inferred utterance-level intent is also helpful in refining the slot filling result. For example, once an \texttt{AddToPlaylist} intent representation is learned in {\ThirdCapsule}, the slot filling may capitalize on the inferred intent representation and recognize slots that are otherwise neglected previously.
To achieve this, we propose a re-routing schema for capsule neural networks, which allows high-level features to be actively engaged in the dynamic routing between {\FirstCapsule} and {\SecondCapsule}, which improves the slot filling performance.

To summarize, the contributions of this work are as follows:
\begin{itemize}
    \item Encapsulating the hierarchical relationship among word, slot, and intent in an utterance by a hierarchical capsule neural network structure.
    \item Proposing a dynamic routing schema with re-routing that achieves synergistic effects for joint slot filling and intent detection.
    \item Showing the effectiveness of our model on two real-world datasets, and comparing with existing models as well as commercial NLU services.
\end{itemize}
\section{Approach}
We propose to model the hierarchical relationship among each word, the slot it belongs to, and the intent label of the whole utterance by a hierarchical capsule neural network structure called {\ModelName}. The proposed architecture consists of three types of capsules: 1) {\FirstCapsule} that learn context-aware word representations, 2) {\SecondCapsule} that categorize words by their slot types via dynamic routing, and construct a representation for each type of slot by aggregating words that belong to the slot, 3) {\ThirdCapsule} determine the intent label of the utterance based on the slot representation as well as the utterance contexts. Once the intent label has been determined by {\ThirdCapsule}, the inferred utterance-level intent helps re-recognizing slots from the utterance by a re-routing schema. 

\subsection{WordCaps}
Given an input utterance \(x = \left (  \mathbf{w}_{1}, \mathbf{w}_{2}, ..., \mathbf{w}_{T}\right )\) of $T$ words, where each word is initially represented by a vector of dimension $D_W$. Here we simply trained word represenations from scratch. Various neural network structures can be used to learn context-aware word representations. For example, a recurrent neural network such as a bidirectional LSTM \citep{hochreiter1997long} can be applied to learn representations of each word in the utterance:
\begin{equation}
\begin{gathered}
     \vec{\mathbf{h}}_{t} = {\rm LSTM}_{fw} (\mathbf{w}_{t}, \vec{\mathbf{h}}_{t-1}),\\
     {\mathord{\buildrel{\lower3pt\hbox{$\scriptscriptstyle\leftarrow$}}\over {\mathbf{h}}} }_{t} = {\rm LSTM}_{bw} (\mathbf{w}_{t}, {\mathord{\buildrel{\lower3pt\hbox{$\scriptscriptstyle\leftarrow$}}\over {\mathbf{h}}} }_{t+1}).\\
\end{gathered}
\end{equation}
For each word \(\mathbf{w}_t\), we concatenate each forward hidden state \(\vec{\mathbf{h}}_{t}\) obtained from the forward ${\rm LSTM}_{fw}$ with a backward hidden state \({\mathord{\buildrel{\lower3pt\hbox{$\scriptscriptstyle\leftarrow$}}\over 
{\mathbf{h}}} }_{t}\) from ${\rm LSTM}_{bw}$ to obtain a hidden state \(\mathbf{h}_t\). The whole hidden state matrix can be defined as \(\mathbf{H} = \left(\mathbf{h}_{1}, \mathbf{h}_{2}, ..., \mathbf{h}_{T}\right) \in \mathbb{R}^{T \times 2D_{H}}\), where $D_H$ is the number of hidden units in each LSTM. 
In this work, the parameters of WordCaps are trained with the whole model, while sophisticated pre-trained models such as ELMo \citep{peters2018deep} or BERT \citep{devlin2018bert} may also be integrated.
 
\subsection{\SecondCapsule}
Traditionally, the learned hidden state \(\mathbf{h}_t\) for each word \(\mathbf{w}_t\) is used as the logit to predict its slot tag. When $\mathbf{H}$ for all words in the utterance is learned, sequential tagging methods like the linear-chain CRF models the tag dependencies by assigning a transition score for each transition pattern between adjacent tags to ensure the best tag sequence of the utterance from all possible tag sequences.

Instead of doing slot filling via sequential labeling which does not directly consider the dependencies among words, the {\SecondCapsule} learn to recognize slots via dynamic routing. The routing-by-agreement explicitly models the hierarchical relationship between capsules. For example, the routing-by-agreement mechanism send a low-level feature, e.g. a word representation in {\FirstCapsule}, to high-level capsules, e.g. {\SecondCapsule}, only when the word representation has a strong agreement with a slot representation.

The agreement value on a word may vary when being recognized as different slots. For example, the word \texttt{three} may be recognized as a \texttt{party\_size\_number} slot or a \texttt{time} slot.  The {\SecondCapsule} first convert the word representation obtained in {\FirstCapsule} with respect to each slot type. We denote \(\mathbf{p}_{k|t}\) as the resulting prediction vector of the $t$-th word when being recognized as the $k$-th slot:
\begin{equation}
    \mathbf{p}_{k|t} = \sigma(\mathbf{W}_{k}\mathbf{h}_{t}^{T} + \mathbf{b}_{k}),
\end{equation}
where $k\in\{1, 2, ..., K\}$ denotes the slot type and $t\in\{1, 2, ..., T\}$. $\sigma$ is the activation function such as $tanh$. $\mathbf{W}_{k}\in \mathbb{R}^{{D_P}\times{2D_H}}$ and $\mathbf{b}_{k} \in \mathbb{R}^{D_P\times 1}$ are the weight and bias matrix for the $k$-th capsule in {\SecondCapsule}, and ${D_P}$ is the dimension of the prediction vector.

\noindent\textbf{Slot Filling by Dynamic Routing-by-agreement}
We propose to determine the slot type for each word by dynamically route prediction vectors of each word from {\FirstCapsule} to {\SecondCapsule}.
The dynamic routing-by-agreement learns an agreement value \(c_{kt}\) that determines how likely the $t$-th word agrees to be routed to the $k$-th slot capsule.  \(c_{kt}\) is calculated by the dynamic routing-by-agreement algorithm \citep{sabour2017dynamic}, which is briefly recalled in Algorithm \ref{al::routing}.

\begin{algorithm}[ht]
\caption{Dynamic routing-by-agreement}\label{al::routing}
\resizebox{7cm}{!}{
\begin{minipage}{1.15\linewidth}
\begin{algorithmic}[1]
\Procedure{Dynamic Routing}{$\mathbf{p}_{k|t}$, $iter$}
    \State {for each {\FirstCapsule} t and {\SecondCapsule} k: ${b}_{kt}\leftarrow0$.}
    
    \For {$iter$~iterations}
        \State {for all {\FirstCapsule} $t$: $\mathbf{c}_t\leftarrow\operatorname{softmax}(\mathbf{b}_t)$}
        
        \State {for all {\SecondCapsule} k: $\mathbf{s}_k \leftarrow \Sigma_r c_{kt}\mathbf{p}_{k|t}$}
        
        \State {for all {\SecondCapsule} k: $\mathbf{v}_k= \operatorname{squash}(\mathbf{s}_{k})$}
        
        \State {for all {\FirstCapsule} t and {\SecondCapsule} k: ${b}_{kt} \leftarrow \text{b}_{kt} + \mathbf{p}_{k|t}\cdot\mathbf{v}_{k}$}
        
	    \EndFor
	\State Return $\mathbf{v}_k$
\EndProcedure
\end{algorithmic}
\end{minipage}
}
\end{algorithm}

The above algorithm determines the agreement value $c_{kt}$ between {\FirstCapsule} and {\SecondCapsule} while learning the slot representations $\mathbf{v}_k$ in an unsupervised, iterative fashion. $\mathbf{c}_t$ is a vector that consists of all $c_{kt}$ where $k\in{K}$.
${b}_{kt}$ is the logit (initialized as zero) representing the log prior probability that the $t$-th word in {\FirstCapsule} agrees to be routed to the $k$-th slot capsule in {\SecondCapsule} (Line 2). During each iteration (Line 3), each slot representation $\mathbf{v}_k$ is calculated by aggregating all the prediction vectors for that slot type $\{{\mathbf{p}_{k|t}} | t{\in}T\}$, weighted by the agreement values $c_{kt}$ obtained from $b_{kt}$ (Line 5-6):
\begin{equation}
    \mathbf{s}_{k} = \sum_{t}^{T}c_{kt}\mathbf{p}_{k|t},
\end{equation}
\begin{equation}
    \mathbf{v}_{k} = \operatorname{squash}(\mathbf{s}_{k}) = \frac{\left \| \mathbf{s}_{k} \right \|^{2}}{1 + \left \| \mathbf{s}_{k}\right \|^{2} }\frac{\mathbf{s}_{k}}{\left \| \mathbf{s}_{k} \right \|},
\end{equation}
where a squashing function $\operatorname{squash}(\cdot)$ is applied on the weighted sum \(\mathbf{s}_{k}\) to get \(\mathbf{v}_{k}\) for each slot type.
Once we updated the slot representation $\mathbf{v}_k$ in the current iteration, the logit $b_{kt}$ becomes larger when the dot product $\mathbf{p}_{k|t}\cdot\mathbf{v}_{k}$ is large. That is, when a prediction vector $\mathbf{p}_{k|t}$ is more similar to a slot representation $\mathbf{v}_k$, the dot product is larger, indicating that it is more likely to route this word to the $k$-th slot type (Line 7). An updated, larger $b_{kt}$ will lead to a larger agreement value $c_{kt}$ between the $t$-th word and the $k$-th slot in the next iteration.
On the other hand, it assigns low $c_{kt}$ when there is inconsistency between $p_{k|t} $ and $\mathbf{v}_k$.
The agreement values learned via the unsupervised, iterative algorithm ensures the outputs of the {\FirstCapsule} get sent to appropriate subsequent {\SecondCapsule} after $iter_{\operatorname{slot}}$ iterations.

\noindent\textbf{Cross Entropy Loss for Slot Filling}\\
For the $t$-th word in an utterance, its slot type is determined as follows:
\begin{equation}
{\hat y}_t = \mathop {\arg \max }\limits_{k \in K} ({{c}_{kt}}).
\end{equation}
The slot filling loss is defined over the utterance as the following cross-entropy function:
\begin{equation}
\mathcal{L}_{slot} = - \sum\limits_t {\sum\limits_k {y_t^k\log (\hat y_t^k)} },
\end{equation}
where $y_t^k$ indicates the ground truth slot type for the $t$-th word. $y_t^k=1$ when the $t$-th word belongs to the $k$-th slot type.

\subsection{\ThirdCapsule}
The {\ThirdCapsule} take the output $\mathbf{v}_k$ for each slot $k \in \{1,2,...,K\}$ in {\SecondCapsule} as the input, and determine the utterance-level intent of the whole utterance.
The {\ThirdCapsule} also convert each slot representation in {\SecondCapsule} with respect to the intent type:
\begin{equation}
\mathbf{q}_{l|k} = \sigma(\mathbf{W}_{l}\mathbf{v}_k^T +b_{l}),
\end{equation}
where $l \in \{1,2,...,L\}$ and $L$ is the number of intents. $\mathbf{W}_l \in \mathbb{R}^{D_L\times D_P}$ and $\mathbf{b}_l \in \mathbb{R}^{D_L\times 1}$ are the weight and bias matrix for the $l$-th capsule in {\ThirdCapsule}.

{\ThirdCapsule} adopt the same dynamic routing-by-agreement algorithm, where:
\begin{equation}
\mathbf{u}_l = \textsc{Dynamic Routing}(\mathbf{q}_{l|k}, iter_{\operatorname{intent}}).
\end{equation}

\noindent\textbf{Max-margin Loss for Intent Detection}\\
Based on the capsule theory, the orientation of the activation vector $\mathbf{u}_l$ represents intent properties while its length indicates the activation probability. 
The loss function considers a max-margin loss on each labeled utterance:
\begin{align}\label{eq:loss}
\begin{split}
\mathcal{L}_{intent}&= \sum_{l=1}^{L}\{
\left[\kern-0.15em\left[ {{z} = z_l} \right]\kern-0.15em\right]\cdot \max (0,m^ +  - {{\left\| {{\mathbf{u}_l}} \right\|}})^2\\
&+ \lambda \left[\kern-0.15em\left[ {{z} \ne z_l} \right]\kern-0.15em\right]\cdot \max (0,{{\left\| {{\mathbf{u}_l}} \right\|}} - m^ - )^2\},\\
\end{split}
\end{align}
where $\|\mathbf{u}_l\|$ is the norm of $\mathbf{u}_l$ and $\left[\kern-0.15em\left[  \right]\kern-0.15em\right]$ is an indicator function, $z$ is the ground truth intent label for the utterance $x$. $\lambda$ is the weighting coefficient, and $m^ +$ and $m^ -$ are margins.

The intent of the utterance can be easily determined by choosing the activation vector with the largest norm ${{\hat z}} = \mathop {\arg \max }\limits_{l \in \{ 1,2,...,L\} } \left\| {{{\mathbf{u}}_l}} \right\|$.

\subsection{Re-Routing}
The {\ThirdCapsule} not only determine the intent of the utterance by the length of the activation vector, but also learn discriminative intent representations of the utterance by the orientations of the activation vectors.
Previously, the dynamic routing-by-agreement shows how low-level features such as slots help construct high-level ideas such as intents. While the high-level features also work as a guide that helps learn low-level features. For example, the \texttt{AddToPlaylist} intent activation vector in {\ThirdCapsule} also helps strength the existing slots such as \texttt{artist\_name} during slot filling on the words \texttt{Sungmin} in {\SecondCapsule}.

Thus we propose a re-routing schema for {\SecondCapsule} where the dynamic routing-by-agreement is realized by the following equation that replaces the Line 7 in Algorithm \ref{al::routing}:
\begin{equation}
\text{b}_{kt} \leftarrow \text{b}_{kt} + \mathbf{p}_{k|t}\cdot\mathbf{v}_{k} + \alpha \cdot \mathbf{p}_{k|t}^{T}{\mathbf{W}_{RR}}\mathbf{\hat u}_{\hat z}^{T},
\end{equation}
where $\mathbf{\hat u}_{\hat z}$ is the intent activation vector with the largest norm.
$\mathbf{W}_{RR}\in\mathbb{R}^{D_P \times D_L}$ is a bi-linear weight matrix, and $\alpha$ as the coefficient.
The routing information for each word is updated toward the direction where the prediction vector not only coincides with representative slots, but also towards the most-likely intent of the utterance.
As a result, the re-routing makes {\SecondCapsule} obtain updated routing information as well as updated slot representations.
\section{Experiment Setup}
To demonstrate the effectiveness of our proposed models, we compare the proposed model {\ModelName} with existing alternatives, as well as commercial natural language understanding services.

\noindent\textbf{Datasets}
For each task, we evaluate our proposed models by applying it on two real-word datasets: SNIPS Natural Language Understanding benchmark\footnote{\url{https://github.com/snipsco/nlu-benchmark/}} (SNIPS-NLU) and the Airline Travel Information Systems (ATIS) dataset \cite{tur2010left}. The statistical information on two datasets are shown in Table \ref{data_statistics}.

\begin{table}[h!]
\centering
\resizebox{7cm}{!}{
\begin{tabular}{l|l|l}
\hline
\textbf{Dataset} & \textbf{SNIPS-NLU} & \textbf{ATIS} \\\hline   
Vocab Size   &  11,241 & 722\\
Average Sentence Length   &  9.05    & 11.28\\
\#Intents  &  7   & 21\\
\#Slots  &  72   & 120\\
\#Training Samples   &  13,084    &  4,478\\
\#Validation Samples   &  700    &  500\\
\#Test Samples   &  700    &  893\\\hline
\end{tabular}
}
\vspace{0.05in}
\caption{Dataset statistics.}
\label{data_statistics}
\end{table}

SNIPS-NLU contains natural language corpus collected in a crowdsourced fashion to benchmark the performance of voice assistants.
ATIS is a widely used dataset in spoken language understanding, where audio recordings of people making flight reservations are collected.

\noindent\textbf{Baselines}
We compare the proposed capsule-based model {\ModelName} with other alternatives:
1) CNN TriCRF \citep{xu2013convolutional} introduces a Convolution Neural Network (CNN) based sequential labeling model for slot filling. The hidden states for each word are summed up to predict the utterance intent. We adopt the performance with lexical features.
2) Joint Seq. \citep{hakkani2016multi} adopts a Recurrent Neural Network (RNN) for slot filling and the last hidden state of the RNN is used to predict the utterance intent. 
3) Attention BiRNN \citep{liu2016attention} further introduces a RNN based encoder-decoder model for joint slot filling and intent detection. An attention weighted sum of all encoded hidden states is used to predict the utterance intent.
4) Slot-gated Full Atten. \citep{goo2018slot} utilizes a slot-gated mechanism as a special gate function in Long Short-term Memory Network (LSTM) to improve slot filling by the learned intent context vector. The intent context vector is used for intent detection.
5) DR-AGG \citep{gong2018information} aggregates word-level information for text classification via dynamic routing. The high-level capsules after routing are concatenated, followed by a multi-layer perceptron layer that predicts the utterance label. We used this capsule-based text classification model for intent detection only.
6) IntentCapsNet \citep{xia2018zero} adopts a multi-head self-attention to extract intermediate semantic features from the utterances, and uses dynamic routing to aggregate semantic features into intent representations for intent detection. We use this capsule-based model for intent detection only.

We also compare our proposed model {\ModelName} with existing commercial natural language understanding services, including api.ai (Now called DialogFlow)\footnote{\url{https://dialogflow.com/}}, Waston Assistant\footnote{\url{https://www.ibm.com/cloud/watson-assistant/}}, Luis\footnote{\url{https://www.luis.ai/}}, wit.ai\footnote{\url{https://wit.ai/}}, snips.ai\footnote{\url{https://snips.ai/}}, recast.ai\footnote{\url{https://recast.ai/}}, and Amazon Lex\footnote{\url{https://aws.amazon.com/lex/}}.

\noindent\textbf{Implementation Details}
The hyperparameters used for experiments are shown in Table \ref{hyperparameter}.

\begin{table}[ht!]
\centering
\resizebox{1.0\linewidth}{!}{
\begin{tabular}{l|llllll}
\hline
\textbf{Dataset}    & $D_W$ & $D_H$ & $D_P$ & $D_L$  &  $iter_{\operatorname{slot}}$ &  $iter_{\operatorname{intent}}$ \\ \hline
SNIPS-NLU &  1024   & 512   &  512 &  128 &   2   &   2    \\ 
ATIS &   1024 &  512  & 512   &  256 &  3  &    3    \\ \hline
\end{tabular}
}
\vspace{0.1in}
\caption{Hyperparameter settings.}
\label{hyperparameter}
\end{table}

\begin{table*}[tb!]
\centering
\resizebox{1.0\textwidth}{!}{
\begin{tabular}{l|ccc|ccc}
\hline
\multirow{2}{*}{\textbf{Model}} & \multicolumn{3}{c|}{\textbf{SNIPS-NLU}} & \multicolumn{3}{c}{\textbf{ATIS}} \\ \cline{2-7} 
                      & Slot (F1) &Intent (Acc) & Overall (Acc) 
                      & Slot (F1) &Intent (Acc) & Overall (Acc)           \\ \hline
CNN TriCRF \cite{xu2013convolutional} 	&-		&-		&-	&0.944   &-	&-	\\
Joint Seq. \cite{hakkani2016multi}  	&0.873	&0.969	&0.732	&0.942   &0.926	&0.807	\\
Attention BiRNN \cite{liu2016attention} &0.878	&0.967	&0.741	&0.942   &0.911	&0.789	\\
Slot-Gated Full Atten. \cite{goo2018slot}	&0.888	&0.970	&0.755	&0.948   &0.936	&0.822	\\
DR-AGG \cite{gong2018information}       & -		&0.966  	& -		& -   	 &0.914 & -	\\
IntentCapsNet \cite{xia2018zero}  		& -		&0.974	& -		& -   	 &0.948 & -	\\
\hline
{\ModelName} &\textbf{0.918}&0.973 &\textbf{0.809}   &\textbf{0.952} &\textbf{0.950} & \textbf{0.834}\\
{\ModelName} w/o Intent Detection &0.902	& -	& -	&{0.948} &-	& -	\\ 
{\ModelName} w/o Joint Training     &0.902	& \textbf{0.977}	&0.804	&0.948  &0.847	&0.743\\ 
\hline
\end{tabular}
}
\vspace{0.15in}
\caption{Slot filling and intention detection results using {\ModelName} on two datasets.}
\label{tab::overall}
\end{table*}
\begin{figure*}[tb!]
    \centering
    \includegraphics[width=\linewidth]{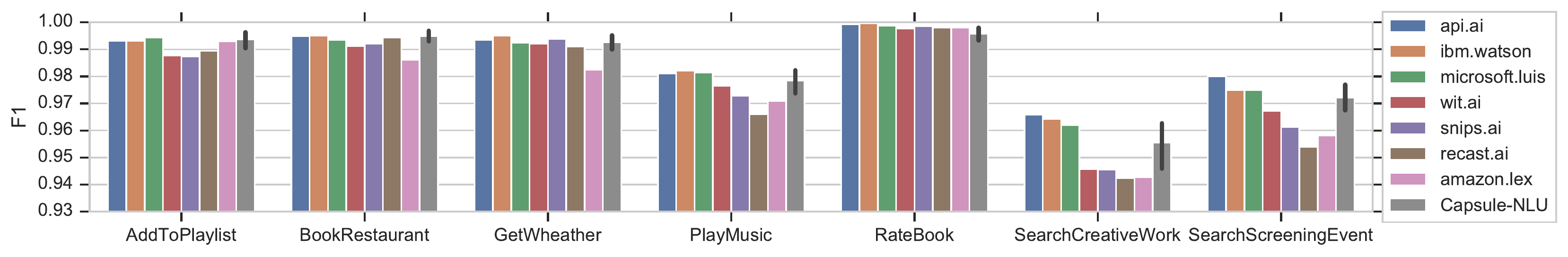}
    \vspace{0.05in}
    \caption{Stratified 5-fold cross validation for benchmarking with existing NLU services on SNIPS-NLU dataset. Black bars indicate the standard deviation.}\label{fig::benchmark}
\end{figure*}

We use the validation data to choose hyperparameters.
For both datasets, we randomly initialize word embeddings using Xavier initializer and let them train with the model.
In the loss function, the down-weighting coefficient $\lambda$ is 0.5, margins $m^ +$ and $m^ -$ are set to 0.8 and 0.2 for all the existing intents. $\alpha$ is set as 0.1.
RMSProp optimizer \citep{tieleman2012lecture} is used to minimize the loss.  To alleviate over-fitting, we add the dropout to the LSTM layer with a  dropout rate of 0.2.

\section{Results}
\noindent\textbf{Quantitative Evaluation} The intent detection results on two datasets are reported in Table \ref{tab::overall}, where the proposed capsule-based model performs consistently better than current learning schemes for joint slot filling and intent detection, as well as capsule-based neural network models that only focuses on intent detection. These results demonstrate the novelty of the proposed capsule-based model {\ModelName} in jointly modeling the hierarchical relationships among words, slots and intents via the dynamic routing between capsules.

Also, we benchmark the intent detection performance of the proposed model with existing natural language understanding services\footnote{https://www.slideshare.net/KonstantinSavenkov/nlu-intent-detection-benchmark-by-intento-august-2017} in Figure \ref{fig::benchmark}.
Since the original data split is not available, we report the results with stratified 5-fold cross validation. From Figure \ref{fig::benchmark} we can see that the proposed model {\ModelName} is highly competitive with off-the-shelf systems that are available to use. Note that, our model archieves the performance without using pre-trained word representations: the word embeddings are simply trained from scratch.

\noindent\textbf{Ablation Study}
To investigate the effectiveness of {\ModelName} in joint slot filling and intent detection, we also report ablation test results in Table \ref{tab::overall}. ``w/o Intent Detection'' is the model without intent detection: only a dynamic routing is performed between {\FirstCapsule} and {\SecondCapsule} for the slot filling task, where we minimize $\mathcal{L}_{slot}$ during training; ``w/o Joint Training'' adopts a two-stage training where the model is first trained for slot filling by minimizing $\mathcal{L}_{slot}$, and then use the fixed slot representations to train for the intent detection task which minimizes $\mathcal{L}_{intent}$. From the lower part of Table \ref{tab::overall} we can see that by using a capsule-based hierarchical modeling between words and slots, the model {\ModelName} w/o Intent Detection is already able to outperform current alternatives on slot filling that adopt a sequential labeling schema. The joint training of slot filling and intent detection is able to give each subtask further improvements when the model parameters are updated jointly.

\noindent\textbf{Visualizing Agreement Values between Capsule Layers}
Thanks to the dynamic routing-by-agreement schema, the dynamically learned agreement values between different capsule layers naturally reflect how low-level features are collectively aggregated into high-level ones for each input utterance. In this section, we harness the intepretability of the proposed capsule-based model via hierarchical modeling and provide case studies and visualizations.

\begin{figure}[h]
    \centering
    \includegraphics[width=\linewidth]{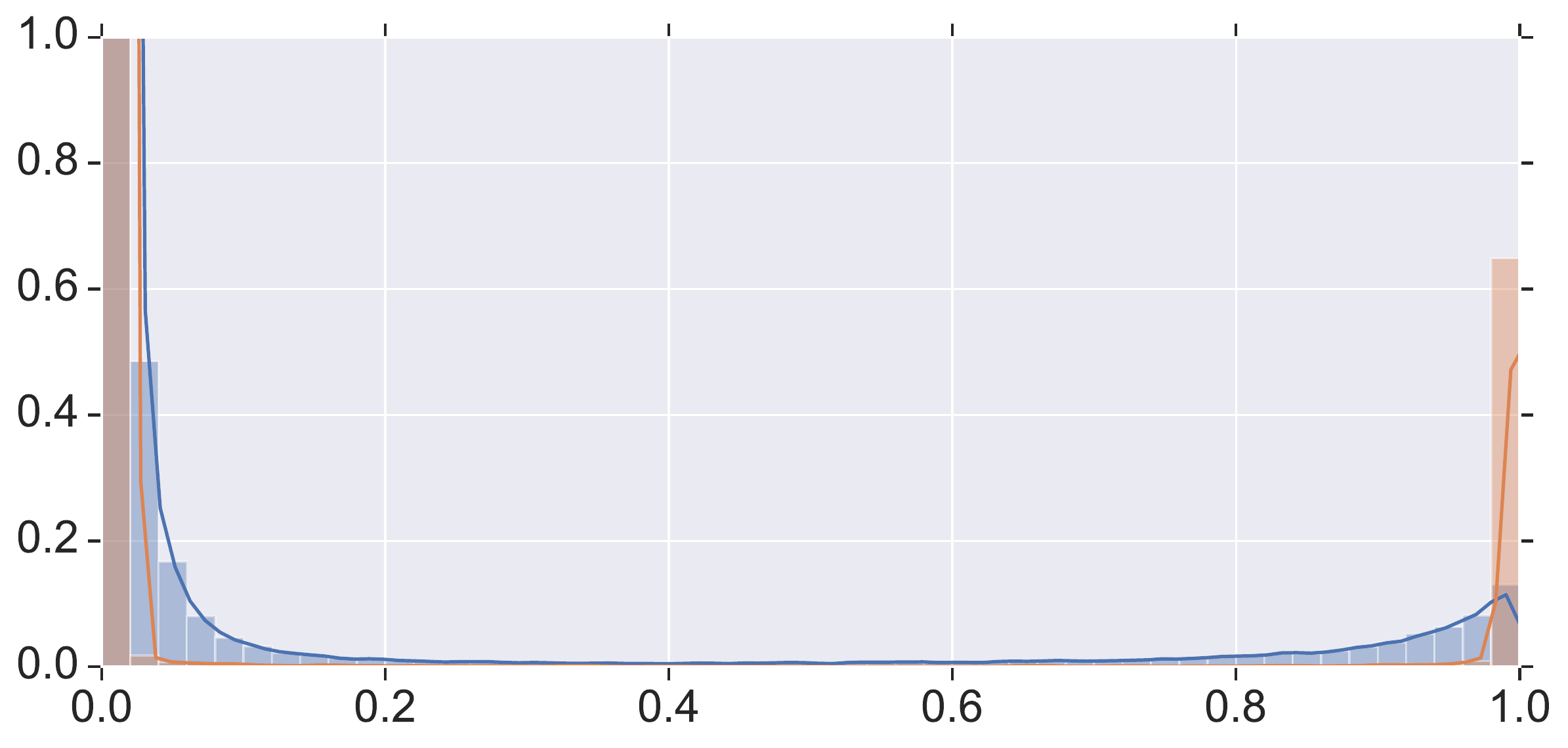}
    \caption{The distribution of all agreement values between WordCaps and SlotCaps on the test split of SNIPS-NLU dataset. Blue: the distribution of values after the first iteration. Yellow: the distribution after the second iteration.}
    \label{fig:w2s_dist}
\end{figure}

\textbf{Between WordCaps and SlotCaps}
First we study the agreement value $c_{kt}$ between the $t$-th word in the WordCaps and the $k$-th slot capsule in SlotCaps.
As shown in Figure \ref{fig:w2s_dist}, we observe that the dynamic routing-by-agreement is able to converge to an agreement quickly after the first iteration (shown in blue bars). It is able to assign a confident probability assignment close to 0 or 1. After the second iteration (shown in orange bars), the model is more certain about the routing decisions: probabilities are more leaning towards 0 or 1 as the model is confident about routing a word in WordCaps to its most appropriate slot in SlotCaps.

\begin{figure}[t!]
    \centering
    \includegraphics[width=\linewidth]{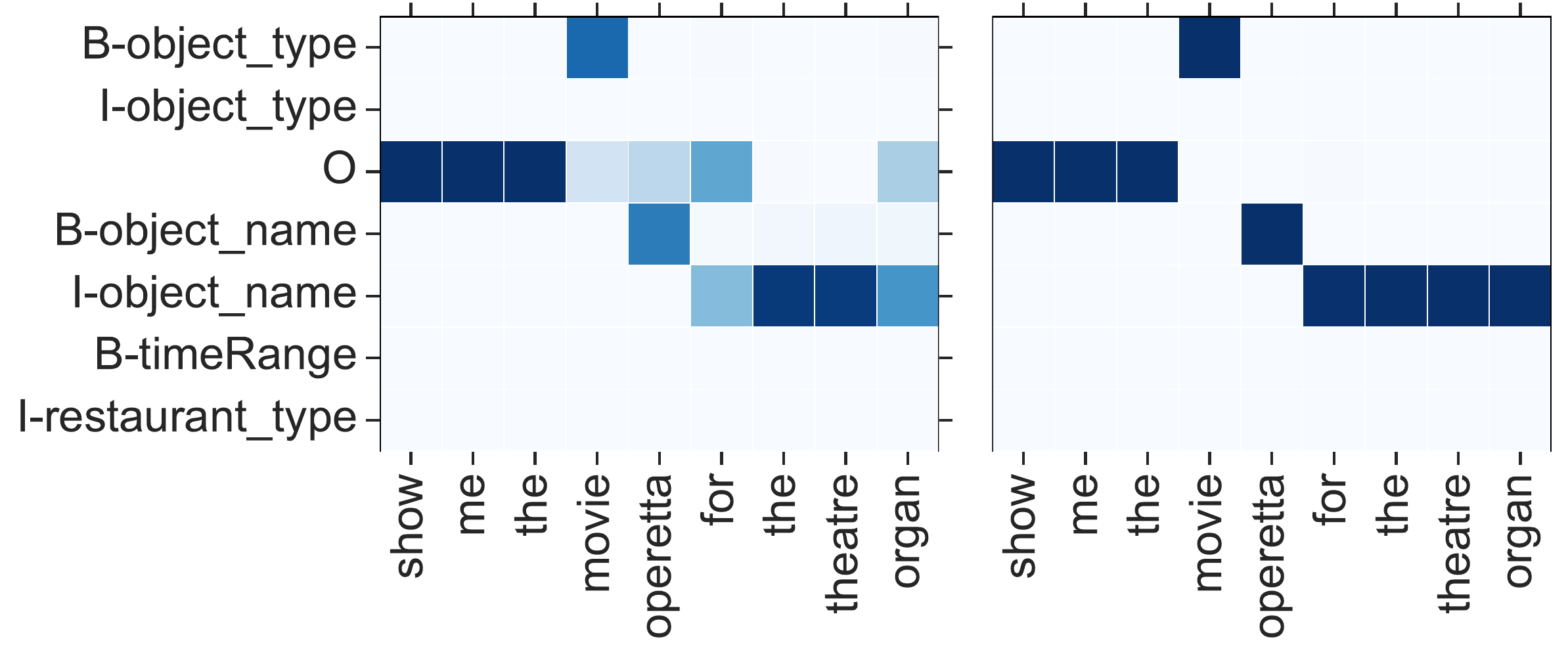}
    \caption{The learned agreement values between WordCaps (x-axis) and SlotCaps (y-axis). A sample from the test split of SNIPS-NLU dataset is shown (Left: after the fist routing iteration. Right: after the second iteration). Due to space limitations, only part of slots (7/72) are shown on the y-axis.}
    \label{fig:w2s}
\end{figure}

However, we do find that when unseen slot values like new object names emerge in utterances like \texttt{show me the movie operetta for the theatre organ} with an intent of \texttt{SearchCreativeWork}, the iterative dynamic routing process would be even more appealing.
Figure \ref{fig:w2s} shows the agreement values learned by dynamic routing-by-agreement. Since the dynamic routing-by-agreement is an iterative process controlled by the variable $iter_{slot}$, we show the agreement values after the first iteration in the left part of Figure \ref{fig:w2s}, and the values after the second iteration in the right part.
 
\begin{figure}[t!]
    \centering
    \includegraphics[width=\linewidth]{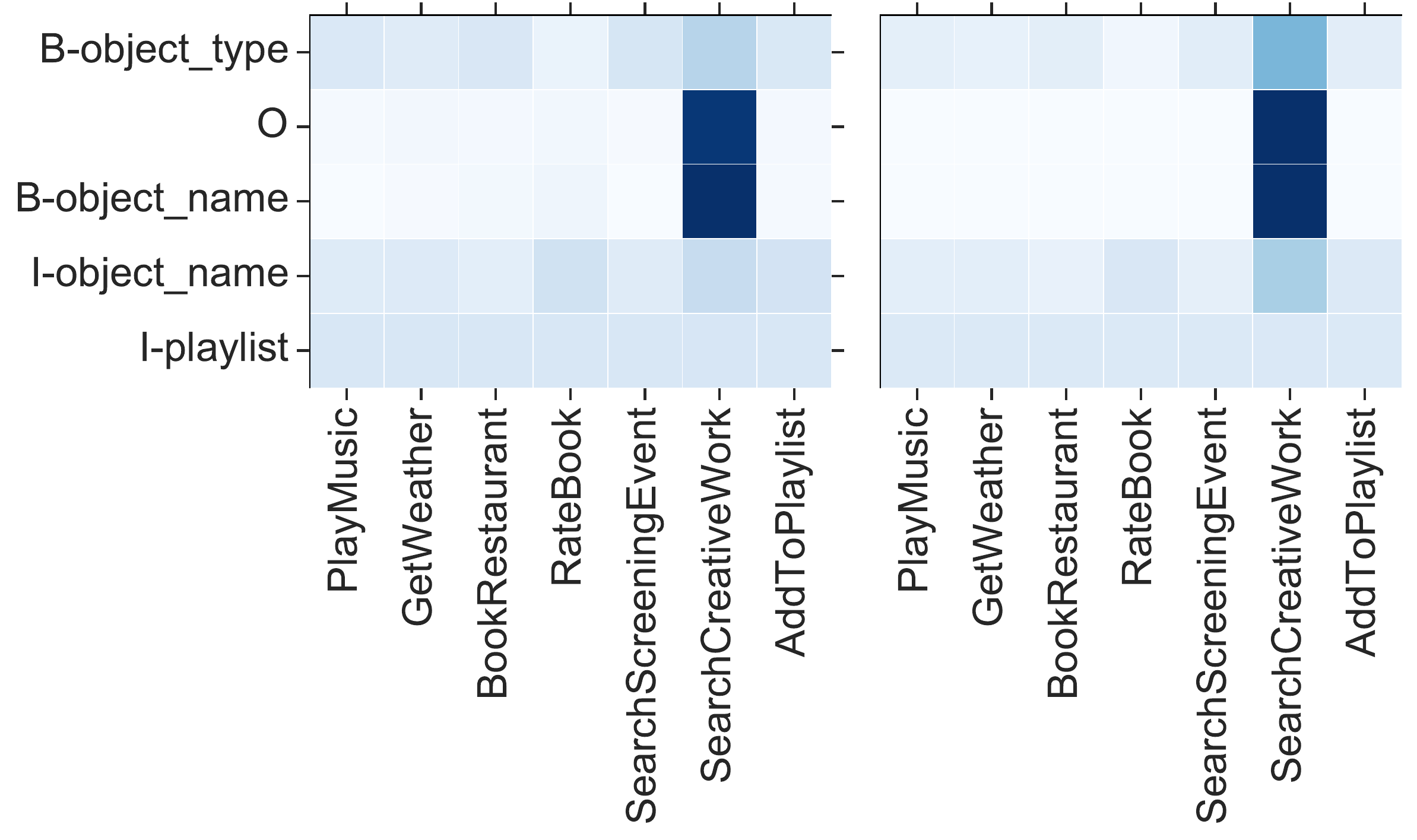}
    \caption{The learned agreement values between SlotCaps (y-axis) and IntentCaps (x-axis). Left: after the first iteration. Right: after the second iteration. The same sample utterance used in Figure 5 is used here.}
    \label{fig:s2i}
    \vspace{-0.3in}
\end{figure}

From the left part of Figure \ref{fig:w2s}, we can see that after the first iteration, the model considers the word \texttt{operetta} itself alone is likely to be an object name, probably because the following word \texttt{for} is usually a context word being annotated as \texttt{O}. Thus it tends to route word \texttt{for} to both the slot \texttt{O} and the slot \texttt{I-object\_name}. However, from the right part of Figure \ref{fig:w2s} we can see that after the second iteration, the dynamic routing found an agreement and is more certain to have \texttt{operetta for the theatre organ} as a whole for the slot \texttt{B-object\_name} and \texttt{I-object\_name}.

\textbf{Between SlotCaps and IntentCaps} Similarly, we visualize the agreement values between each slot capsule in SlotCaps and each intent capsule in IntentCaps.
The left part of Figure \ref{fig:s2i} shows that after the first iteration, since the model is not able to correctly recognize \texttt{operetta for the theatre organ} as a whole, only the context slot \texttt{O} (correspond to the word \texttt{show me the}) and \texttt{B-object\_name} (correspond to the word \texttt{operetta}) contribute significantly to the final intent capsule. From the right part of Figure \ref{fig:s2i}, we found that with the word  \texttt{operetta for the theatre organ} being recognized in the lower capsule, the slots \texttt{I-object\_name} and \texttt{B-object\_type} contribute more to the correct intent capsule \texttt{SearchCreativeWork}, when comparing with other routing alternatives to other intent capsules.
\section{Related Works}
\noindent\textbf{Intent Detection}
With recent developments in deep neural networks, user intent detection models \citep{hu2009understanding,xu2013convolutional,zhang2016mining,liu2016attention,zhang2017bringing,chen2016end,xia2018zero} are proposed to classify user intents given their diversely expressed utterances in the natural language. As a text classification task, the decent performance on utterance-level intent detection usually relies on hidden representations that are learned in the intermediate layers via multiple non-linear transformations. 

Recently, various capsule based text classification models are proposed that aggregate word-level features for utterance-level classification via dynamic routing-by-agreement \citep{gong2018information,zhao2018investigating,xia2018zero}. Among them, \citet{xia2018zero} adopts self-attention to extract intermediate semantic features and uses a capsule-based neural network for intent detection. 
However, existing works do not study word-level supervisions for the slot filling task. In this work, we explicitly model the hierarchical relationship between words and slots on the word-level, as well as intents on the utterance-level via dynamic routing-by-agreement.

\noindent\textbf{Slot Filling}
Slot filling annotates the utterance with finer granularity: it associates certain parts of the utterance, usually named entities, with pre-defined slot tags. Currently, the slot filling is usually treated as a sequential labeling task. A recurrent neural network such as Gated Recurrent Unit (GRU) or Long Short-term Memory Network (LSTM) is used to learn context-aware word representations, and Conditional Random Fields (CRF) are used to annotate each word based on its slot type. Recently, \citet{shen2017disan, tan2017deep} introduce the self-attention mechanism for CRF-free sequential labeling.

\noindent\textbf{Joint Modeling via Sequence Labeling}
To overcome the error propagation in the word-level slot filling task and the utterance-level intent detection task in a pipeline, joint models are proposed to solve two tasks simultaneously in a unified framework.
\citet{xu2013convolutional} propose a Convolution Neural Network (CNN) based sequential labeling model for slot filling. The hidden states corresponding to each word are summed up in a classification module to predict the utterance intent. A Conditional Random Field module ensures the best slot tag sequence of the utterance from all possible tag sequences.
\citet{hakkani2016multi} adopt a Recurrent Neural Network (RNN) for slot filling and the last hidden state of the RNN is used to predict the utterance intent. 
\citet{liu2016attention} further introduce an RNN based encoder-decoder model for joint slot filling and intent detection. An attention weighted sum of all encoded hidden states is used to predict the utterance intent.
Some specific mechanisms are designed for RNNs to explicitly encode the slot from the utterance. For example, \citet{goo2018slot} utilize a slot-gated mechanism as a special gate function in Long Short-term Memory Network (LSTM) to improve slot filling by the learned intent context vector.
However, as the sequence becomes longer, it is risky to simply rely on the gate function to sequentially summarize and compress all slots and context information in a single vector \cite{cheng2016long}.

In this paper, we harness the capsule neural network to learn a hierarchy of feature detectors and explicitly model the hierarchical relationships among word-level slots and utterance-level intent. Also, instead of doing sequence labeling for slot filling, we use a dynamic routing-by-agreement schema between capsule layers to route each word in the utterance to its most appropriate slot type. And we further route slot representations, which are learned dynamically from words, to the most appropriate intent capsule for intent detection.
\section{Conclusions}
In this paper, a capsule-based model, namely {\ModelName}, is introduced to harness the hierarchical relationships among words, slots, and intents in the utterance for joint slot filling and intent detection.
Unlike treating slot filling as a sequential prediction problem, the proposed model assigns each word to its most appropriate slots in {\SecondCapsule} by a dynamic routing-by-agreement schema. The learned word-level slot representations are futher aggregated to get the utterance-level intent representations via dynamic routing-by-agreement. A re-routing schema is proposed to further synergize the slot filling performance using the inferred intent representation.
Experiments on two real-world datasets show the effectiveness of the proposed models when compared with other alternatives as well as existing NLU services.
\section{Acknowledgments}
We thank the reviewers for their valuable comments. This work is supported in part by NSF through grants IIS-1526499, IIS-1763325, and CNS-1626432.
\bibliography{ref.bib}
\bibliographystyle{acl_natbib}
\balance
\end{document}